\documentclass{article}
\usepackage{amsmath,graphicx, url}
\usepackage{amssymb, bm}
\usepackage{pifont}
\newcommand{\cmark}{\ding{51}}%
\newcommand{\xmark}{\ding{55}}%

\usepackage{INTERSPEECH2021}
\title{On Minimum Word Error Rate Training of the Hybrid \\ Autoregressive Transducer}
\name{Liang Lu, Zhong Meng, Naoyuki Kanda, Jinyu Li, and Yifan Gong}
\address{Microsoft Corp., USA}
\email{\{liang.lu, nakanda, zhong.meng, jinyli, yifan.gong\}@microsoft.com}
\begin{document}

\maketitle
\begin{abstract}
Hybrid Autoregressive Transducer (HAT) is a recently proposed end-to-end acoustic model that extends the standard Recurrent Neural Network Transducer (RNN-T) for the purpose of the external language model (LM) fusion. In HAT, the blank probability and the label probability are estimated using two separate probability distributions, which provides a more accurate solution for internal LM score estimation, and thus works better when combining with an external LM. Previous work mainly focuses on HAT model training with the negative log-likelihood loss, while in this paper, we study the minimum word error rate (MWER) training of HAT -- a criterion that is closer to the evaluation metric for speech recognition, and has been successfully applied to other types of end-to-end models such as sequence-to-sequence (S2S) and RNN-T models. From experiments with around 30,000 hours of training data, we show that MWER training can improve the accuracy of HAT models, while at the same time, improving the robustness of the model against the decoding hyper-parameters such as length normalization and decoding beam during inference. 
\end{abstract}
\noindent\textbf{Index Terms}: Hybrid autoregressive transducer, Minimum word error rate loss, End-to-end speech recognition  

\section{Introduction}
\label{sec:intro}

End-to-end (E2E) speech recognition has been an active research topic for the past few years, and there has been a tremendous progress in this field. In particular, two categories of E2E models have been explored for speech recognition tasks, i.e., label-synchronous model such as sequence-to-sequence with attention model (S2S)~\cite{chorowski2015attention, lu2015study, chan2016listen}, and time-synchronous model such as connectionist temporal classification (CTC)~\cite{graves2006connectionist} and recurrent neural network transducer (RNN-T)~\cite{graves2012sequence}. Compared to CTC, RNN-T can perform joint acoustic and language modeling with the augmented prediction network, and therefore it usually yields higher recognition accuracy. Since the time-synchronous model is more natural and straightforward for streaming speech recognition, most recent research efforts have been focusing on RNN-T and its variants, and the model has been successfully applied for on-device speech recognition~\cite{he2019streaming}. 

The conventional hybrid speech recognition system integrates 
individual acoustic model and language model
(LM) by the Bayes rule during decoding. It usually requires a large decoding graph for large-vocabulary speech recognition problems, and therefore is not very feasible to be deployed to computation and memory constrained platforms such as mobile devices. In contrast, the E2E approach folds the acoustic model, lexicon and LM component into a single model, which enjoys the advantage of a smaller memory footprint. This feature makes E2E models more preferable for on-device speech recognition scenarios with very limited computational capacity and memory. However, by performing joint acoustic and language modeling, E2E models have a critical drawback that they loose the flexibility for target-domain LM adaption. Indeed, external LM fusion has been a long lasting research problem since the emergence of the E2E approach in speech recognition, and numerous approaches have been proposed to tackle this problem~\cite{Hannun2014Deep, gulcehre2015using, chorowski2016towards, kanda2016maximum, sriram2017cold, kanda2017maximum}.  More recently, McDermott et al.~\cite{mcdermott2019density} proposed the density ratio approach for LM fusion that demonstrates higher accuracy compared to the popular shallow fusion approach. 

Instead of LM fusion, another approach is to modularize the E2E models so that the acoustic model and LM factor can be more disentangled. Along this line, the hybrid autoregressive transducer (HAT)~\cite{variani2020hybrid}  builds upon the standard RNN-T model by introducing two distinct distributions for labels and the blank token respectively. The advantage is that the internal LM score is readily available by ignoring the blank probabilities, which can be used for the external LM fusion by a scoring function similar to the one used in the density ratio approach. In~\cite{variani2020hybrid}, the authors mainly investigated HAT with the negative log-likelihood (NLL) training criterion. In this paper, we study the minimum word error rate (MWER) training criterion for HAT.  While there have been a few studies of MWER for E2E models such as S2S~\cite{prabhavalkar2018minimum, weng2018improving} and RNN-T~\cite{weng2019minimum, guo2020efficient}, this work concerns the HAT model, which may be viewed as a variant of RNN-T, and we discuss the robustness of MWER training against the decoding hyperparameters such as length normalization and the decoding beam that has not been well covered in the existing literature. The contributions of the paper are summarized as follows.
\begin{enumerate}
    \item We study the MWER training of HAT models, and discuss the robustness of MWER training against length normalization and the size of inference beam.
    \item We perform a comparison of HAT and RNN-T with and without length normalization
    \item We investigate the HAT inference with an external LM with and without MWER training. 
\end{enumerate}

\section{Related work}

Minimum Bayesian Risk (MBR) training has been successfully applied to speech recognition within the hybrid framework~\cite{povey2002minimum, gibson2006hypothesis, kingsbury2012scalable, Vesely:IS13} with a long history, which can operate at different granularities such as. phone-level MPE~\cite{povey2002minimum} or state-level sMBR~\cite{gibson2006hypothesis, kingsbury2012scalable}. Word-level MBR, which is also referred to as MWER, is firstly proposed for the hybrid model~\cite{shannon2017optimizing}, but the same type of criterion was also shown to improve end-to-end models such as sequence-to-sequence model~\cite{prabhavalkar2018minimum, weng2018improving} and RNN-T~\cite{weng2019minimum, guo2020efficient}. Since the exact sequence-level posterior is computationally intractable, Shannon~\cite{shannon2017optimizing} proposed a sampling approach, while in~\cite{prabhavalkar2018minimum}, the authors showed that computing the posterior from an N-best list worked equally well or even better with a much simpler implementation. As for RNN-T models, a single alignment of each hypothesis was used to derive the gradients of the MWER loss~\cite{weng2019minimum}, while in~\cite{guo2020efficient}, the gradients were computed by marginalizing all the possible alignments. In this work, we study the MWER training of HAT, which is a variant of RNN-T model that uses two distributions to estimate the blank token probability and label probability, respectively. We used the N-best list approach to estimate the sequence-level posteriors by marginalizing over all the possible alignments as in~\cite{guo2020efficient}, and to compute the gradients. We study the robustness of MWER training against the decoding hyper-parameters that was briefly discussed in~\cite{peyser2020improving} in the context of speech recognition of rare words with LM fusion. 

\section{RNN-T and HAT}
\label{sec:hat}

\subsection{Recurrent Neural Network Transducer}
\label{ssec:rnnt}

RNN-T performs sequence transduction in a time-synchronous fashion. Given an acoustic feature sequence $\bm{x} = \{x_1, \cdots, x_T\}$ and its corresponding label sequence $\bm{y} = \{y_1, \cdots, y_U\}$, where $T$ is the length of the acoustic sequence, and $U$ is the length of the label sequence, RNN-T defines the conditional probability as
\begin{align}
\label{eq:prob}
P(\bm{y} \mid \bm{x}) = \sum_{\tilde{\bm y} \in \mathcal{B}^{-1}(Y)}{P(\tilde{\bm y} \mid \bm{x})},
\end{align}
where $\tilde{\bm y}$ is a path that contains the blank token $\O$, and the function $\mathcal{B}$ denotes mapping the path to $\bm y$ by removing the blank tokens in $\tilde{\bm y}$. Essentially, the probability $P(\bm{y} \mid \bm{x})$ is calculated by summing over the probabilities of all the possible paths that can be mapped to the label sequence after applying the function $\mathcal{B}$. The probability can be efficiently computed by the forward-backward algorithm, which requires to compute the probability of each step, i.e.,  
\begin{align}
P(k \mid x_{[1:t]}, y_{[1:u]}) = \frac{\exp \left( J(f_t^k + g_u^k) / Z \right)}{\sum_{k^\prime \in \bar{\mathcal{V}}}\exp \left(J(f_t^{k^\prime} + g_u^{k^\prime}) / Z  \right)},
\end{align}
where ${\bm f}_t$ and ${\bm g}_u$ are the output vectors from the audio encoder network and the transcription network at the time step $t$ and $u$ respectively, and $J(\cdot)$ denotes a non-linear function followed by an affine transform. $\bar{\mathcal{V}}$ denotes the set of the vocabulary $\mathcal{V}$ with an additional blank token, i.e., $\bar{\mathcal{V}} = \mathcal{V} \cup \O$. $Z$ is the temperature for the Softmax function, which is usually set to be 1. Given the distribution of each timestep $(t, u)$, the sequence-level conditional probability as Eq. \eqref{eq:prob} can be obtained by the forward-backward algorithm, where the forward variable is defined as
\begin{align*}
\alpha(t, u) & = \alpha(t-1, u) P(\O \mid  x_{[1:t-1]}, y_{[1:u]}) \\ 
 & + \alpha(t, u-1) P(y_u \mid x_{[1:t]}, y_{[1:u-1]}),
\end{align*}
while the backward variable can be defined similarly. The probability $P({\bm y} \mid {\bm x})$ can be computed as
\begin{align}
P(\bm{y} \mid \bm{x}) = \alpha(T, U) P(\O | x_{[1:T]}, y_{[1:U]}).
\end{align}
The model is usually trained by minimizing the negative log-likelihood (NLL) as
\begin{align}
\label{eq:nll}
\mathcal{L_{\text{NLL}}} = -\log P(\bm{y} \mid \bm{x}).
\end{align}
Details about the derivatives of the loss can be found in~\cite{graves2012sequence}. 

\subsection{Hybrid Autoregressive Transducer}
\label{ssec:hat}

The hybrid autoregressive transducer (HAT)~\cite{variani2020hybrid} is proposed as an extension of the standard RNN-T, which aims to disentangle the LM factor and the acoustic model factor, and approximate the internal LM score for external LM fusion similar to the form of the density ratio approach~\cite{mcdermott2019density}. Approximating the internal LM score in the standard RNN-T model is not straightforward, as the token probability is computed over the extended vocabulary set $\bar{\mathcal{V}}$ instead of $\mathcal{V}$, indicating that each label probability is biased even after we remove the acoustic feature $f_t^k$ in Eq.~\eqref{eq:prob}. HAT reformulates the token probability, and uses two distributions for the blank token and label tokens, i.e.,
\begin{align}
\label{eq:prob2}
P(k \mid x_{[1:t]}, y_{[1:u]}) =\left\{
 \begin{array}{ll}
 b_{t,u} & k = \O \\
 (1-b_{t,u})\tilde{P}(k \mid x_{[1:t]}, y_{[1:u]})  & k = y_{u+1}
 \end{array}
 \right.
\end{align}
where $b_{t,u}$ is the blank token probability, which can be drawn from a Bernoulli distribution that can be estimated using a Sigmoid function.  $\tilde{P}(k\mid x_{[1:t]}, y_{[1:u]})$ is the label distribution, which is obtained by a Softmax function over the vocabulary set $\mathcal{V}$ instead of $\bar{\mathcal{V}}$, i.e.,
\begin{align}
\tilde{P}(k \mid x_{[1:t]}, y_{[1:u]}) = \frac{\exp \left( J(f_t^{k} + g_u^{k}) \right)}{\sum_{k^\prime \in \mathcal{V}}\exp \left(J(f_t^{k^\prime} + g_u^{k^\prime}) \right)}.
\end{align}
Given the new probability definition as Eq.~\eqref{eq:prob2}, we can follow the same forward-backward algorithm to compute the sequence-level conditional probability and train the model with NLL loss as Eq.~\eqref{eq:nll}. Since the blank distribution and the label distribution are now disentangled, we can approximate the internal LM score of the HAT model by feeding in the transcription network outputs only to the Softmax function, i.e.,
\begin{align} 
\label{eq:ilm}
\tilde{P}_{ILM}(y) &= \prod_1^U \tilde{P}(y_u \mid  y_{1:u-1}) \\
&= \prod_1^U \frac{\exp \left( J(g_{u-1}^{y_u}) \right)}{\sum_{k^\prime \in \mathcal{V}}\exp \left(J(g_{u-1}^{k^\prime}) \right)}.
\end{align}
During inference, HAT adopts the similar decoding rule as in the density ratio approach~\cite{mcdermott2019density} when the external LM is available. The best path is obtained by maximizing the interpolated score as
\begin{align}
\label{eq:dec1}
\bm{y}^* = &\arg\max_{\bm y} \Big ( \big( \log P(\bm y \mid \bm x) -\lambda_1 \log \tilde{P}_{ILM}(\bm y)  \nonumber   \\ 
&+  \lambda_2 \log P_{LM}(\bm y) \big) / |\bm y| \Big ),
\end{align}
where $\lambda_2$ and $\lambda_2$ are interpolation weights, and $P_{LM}(\bm y)$ denotes the probability from the external LM, which may be trained on a large amount of text-only data, or from a specific domain. The final score is normalized by the length of $\bm y$, denoted as $|\bm y|$, which is referred to as length normalization. As explained in~\cite{graves2012sequence}, without length normalization the decoder tends to prefer short transcription, and thus resulting in high truncation error.  In the case that the external LM is not available, the HAT model can be decoded in the same way as the standard RNN-T model, i.e., 
\begin{align}
\label{eq:dec2}
{\bm y}^* = &\arg\max_{\bm y} \left( \left( \log P({\bm y} \mid \bm x)\right) / |\bm y|\right).
\end{align}
However, the length normalization approach is not favorable to the streaming scenario, especially with the best-first beam search algorithm~\cite{graves2012sequence} as the best path may alter more frequently. We will discuss its impact for both RNN-T and HAT in our experimental section. 

\section{Minimum Word Error Rate Training}
\label{sec:mwer}

The MWER training criterion, which is also referred to as word-level MBR, aims to mitigate the mismatch between the training criterion and the evaluation metric of a speech recognition model. Instead of minimizing the NLL loss as Eq.~\eqref{eq:nll}, MWER minimizes the expected word errors~\cite{shannon2017optimizing}. In the context of RNN-T and HAT, the MWER loss can be formulated as 
\begin{align} 
\mathcal{L}_{\text{MWER}} = \sum_{\bm{y}_i} \hat{P}(\bm{y}_i \mid \bm x) R({\bm y}_i, {\bm y}^r),
\end{align}
where $\hat{P}({\bm y}_i \mid \bm x)$ denotes the posterior probability of an hypothesis ${\bm y}_i$, and $R(\cdot)$ denotes the risk function, which measures the edit-distance between the hypothesis ${\bm y}_i$ and the reference transcription ${\bm y}^r$ at the word-level in MWER. While the exact posterior probability is computationally intractable, in practice, an N-best list of hypotheses from beam search decoding are used to compute the empirical posterior probability~\cite{prabhavalkar2018minimum}, which can be expressed as
\begin{align}
\label{eq:post}
\hat{P}({\bm y}_i \mid \bm x) = \frac{P({\bm y}_i \mid \bm x)}{\sum_{{\bm y}_i} P({\bm y}_i | \bm x)},
\end{align}
where $P({\bm y}_i | \bm x)$ is the conditional probability defined as Eq. \eqref{eq:prob} for RNN-T and HAT. To train the model by gradient descent algorithm, we can compute the derivative of $\mathcal{L}_{\text{MWER}}$ with respect to $\log P({\bm y}_i \mid \bm x)$ instead of the log probability of a specific alignment~\cite{weng2019minimum}, which is the same algorithm used in~\cite{guo2020efficient}, i.e.,
\begin{align}
\label{eq:mwer}
\frac{\partial {\mathcal L}_{\text{MWER}}}{\partial \log P({\bm y}_i \mid \bm x)} = \hat{P}({\bm y}_i \mid \bm x) \left(R({\bm y}_i, {\bm y}^r) - \bar{R}\right),
\end{align}
where $\bar{R}$ denotes the expected risk as $\bar{R} = \sum_{{\bm y}_i} \hat{P}({\bm y}_i \mid \bm x) R({\bm y}_i, {\bm y}^r)$, which is the weighted average word error in the case of MWER. As pointed out in~\cite{shannon2017optimizing, prabhavalkar2018minimum}, intuitively the MWER criterion increases the posterior probability of the hypotheses whose word errors are smaller than the expected risk, while decreases the posterior probabilities of those whose word errors are larger than the expected risk. Given Eq.~\eqref{eq:mwer}, we can compute the derivative of the MWER loss with respect to the probability of each token according to the chain rule, i.e., 
\begin{align}
\frac{\partial {\mathcal L}_{\text{MWER}}}{\partial P(k \mid x_{[1:t]}, y_{i[1:u]})} &= \frac{\partial {\mathcal L}_{\text{MWER}}}{\partial \log P({\bm y}_i \mid \bm x)} \nonumber \\
& \times \frac{\partial \log P({\bm y}_i \mid \bm x)}{\partial P(k \mid x_{[1:t]}, y_{i[1:u]})}.
\end{align}
Detailed information regarding the computation of $\frac{\partial \log P({\bm y}_i \mid \bm x)}{\partial P(k \mid x_{[1:t]}, y_{i[1:u]})}$ can be found in~\cite{graves2012sequence}. 

\section{Experiments and Results}
\label{sec:exp}

In our experiments, the models were trained with around 30,000 hours of anonymized and transcribed Microsoft data, recorded in various conditions. Our evaluation dataset has around 260,000 utterances from various domains including both reading and conversational speech recorded in close-talk and far-filed microphones. We used 4,000 word-piece units including the blank token for tokenization. The RNN-T and HAT model investigated in our experiments share the same neural network architecture, namely, a 6-layer unidirectional LSTM as the audio encoder, and a 2-layer unidirectional LSTM as the transcription network. We used a linear layer followed by the ReLU activation function as the joint network. The number of cell units in both audio encoder and transcription network is 1024. In terms of acoustic features, we used 80-dimensional log-mel filter banks, which are sampled at a 10 millisecond frame rate. We then spliced the consecutive frames with the context window of 8 to generate the super-frames, and then downsampled the acoustic sequences by the factor of 3. 
 
 \begin{table}[t]\centering
\caption{Baseline RNN-T and HAT word error rates (WERs) without an external LM. The decoding beam size was set to 8.}
\label{tab:baseline}
\footnotesize
\vskip0.15cm
\begin{tabular}{l|ccc}
\hline 

\hline
Model   & Temperature    & Length Norm & WER  \\ \hline
  RNN-T & 1.0 & \cmark &  16.6\\
 RNN-T & 1.2 & \cmark &  16.4\\
 RNN-T & 1.2 & \xmark &  16.8\\ \hline
 HAT & 1.0 & \cmark & 16.4 \\
HAT & 1.2 & \cmark & 16.3 \\
HAT & 1.2 & \xmark & 17.2 \\

\hline

\hline
\end{tabular}
\vskip-3mm
\end{table}

\subsection{Baseline Results without External LM}
\label{ssec:baseline}

We first show the baseline results of RNN-T and HAT in terms of word error rate (WER) in Table \ref{tab:baseline}. 
In this experiment, we did not use any external LM, and therefore both RNN-T and HAT models were evaluated following Eq. \eqref{eq:dec2}.
We used the same configuration to train both models, e.g, the same learning rate scheduler, minibatch size, etc. We trained the model using 32 GPUs with data parallelism, and it took around 4 days for the model to converge. 
The results in Table \ref{tab:baseline} show that without the external LM, RNN-T and HAT can achieve very comparable WERs. The average WER of the 26,000 utterances evaluation set is around 16 - 17\%.
Following~\cite{guo2020efficient}, we also evaluated the impact of the temperature, and observed that by increasing the temperature to 1.2 during inference, we can slightly improve the accuracy. 

We also evaluated the models without length normalization during inference. The results are shown in the 3rd and 6th rows of Table \ref{tab:baseline}. which were obtained by applying length normalization at the end of the decoding. In the streaming scenario, it is unclear if applying length normalization at each decoding step will introduce difference in terms of WER, and it may also complicate the decoding path expansion depending on the beam search algorithm been used. One solution could show transcription without length normalization, and when the end of sentence token is detected, the system may update the transcription by applying length normalization. However, drastic change of the transcription may hurt the user experience, and it is desirable that the model is robust to length normalization. From our results, with or without length normalization during inference, the difference in terms of WERs is only around 2\% relative for the standard RNN-T model, while for the HAT model, the difference is over 5\%. The difference may stem from the way HAT estimates the blank probability, which may incur larger variance of the sequence-level probability compared to the standard RNN-T. However, detailed reason requires further investigation. 

\begin{table}[t]\centering
\caption{Comparison of NLL training and MWER training of HAT with/without length normalization during decoding. The temperature in the Softmax was 1.2 during inference, and the decoding beam was 8.}
\label{tab:mwer}
\footnotesize
\vskip0.15cm
\begin{tabular}{l|cc}
\hline 

\hline
Loss       & Length Norm & WER  \\ \hline
NLL  & \cmark & 16.3 \\
 MWER  & \cmark & 15.9 \\ \hline
 NLL  & \xmark & 17.2 \\
MWER  & \xmark & 16.5 \\
\hline

\hline
\end{tabular}
\vskip-3mm
\end{table}

\subsection{MWER Training, Length Normalization and Decoding Beam}

We then study MWER training for the HAT model. We used the model from Table~\ref{tab:baseline} as the seed model, and then trained the model with the MWER criterion as Eq.~\eqref{eq:mwer} using a small fixed learning rate as $1 \times 10^{-6}$. We used the N-best list approach to compute the posteriors as in Eq.~\eqref{eq:post}, for which the N-best list were generated on-the-fly with decoding beam size of 4. For MWER training, we used 32 GPUs, and due to the memory constraint, we set the minibatch size as 1 for each GPU, so for each model update, the effective minibatch size becomes 32. We trained the model with around 6,000 hours of data until the model was fully converged according to the validation loss. We evaluated the MWER trained model with and without length normalization during decoding, and the results are given in Table \ref{tab:mwer}. Without length normalization, MWER training can reduce the WER by around 4\% relative compared with the result of the NLL trained model. The gain is comparable to the results reported in~\cite{guo2020efficient} with an RNN-T model, which did not apply length normalization during decoding either. However, the gain after length normalization during inference is slightly smaller (15.9 vs. 16.3), although MWER can improve the robustness of the model against this decoding hyperparameter.  

\begin{figure}[t]
\small
\centerline{\includegraphics[width=0.45\textwidth]{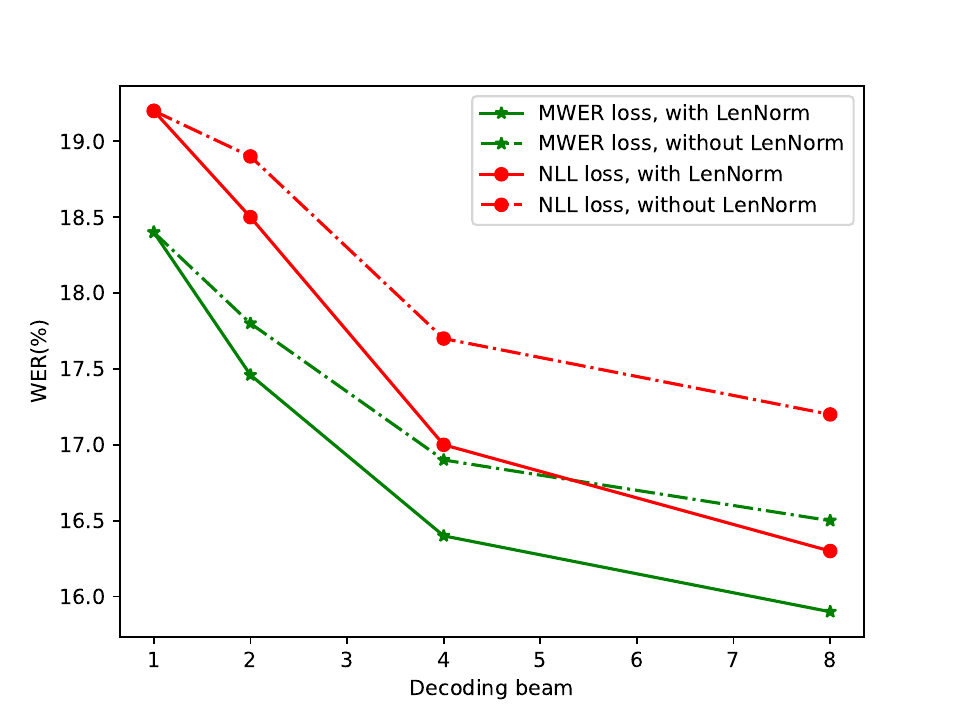}}
\vskip-4mm
\caption{Comparison of the models trained with NLL loss and MWER loss using different decoding beams. The dashed lines correspond to the results without length normalization (LenNorm) during inference, while the solid lines corresponds to those with LenNorm.}  
\label{fig:beam}
\vskip -5mm
\end{figure}

We also evaluated the models with difference decoding beams to study the robust of the model against this decoding hyperparameter. The results are shown in Figure \ref{fig:beam}. We observe that the model trained with the NLL loss is more sensitive to the decoding beam, i.e., the WER increases rapidly when the decoding beam goes smaller from 8. In contrast, the MWER trained model is more robust against the decoding beam size. In addition, the gap between decoding with or without length normalization is much smaller compared with the NLL trained model. Overall, our results demonstrate that MWER training can improve the robustness of the model against the decoding hyperparmaters, which is a desirable  feature in real applications.

\subsection{LM Fusion}
\label{sec:lm}

As discussed before, HAT is a more modularized E2E model compared with the standard RNN-T in the sense that the acoustic and LM factors can be approximately disentangled, enabling more efficient and accurate LM fusion. To study this aspect of HAT in the context of MWER training, we performed a set of experiments with an external LM. Our LM is based on RNNs, which has two LSTM layers, and each layer has 2048 hidden units. The model was trained on text including short message dictation and conversational data, which has around 2 billion words in total. We used the same word-piece tokenization as used for the HAT model training. Following~\cite{variani2020hybrid}, we computed the internal LM scores according to Eq.~\eqref{eq:ilm} and performed decoding according to Eq.~\eqref{eq:dec1}. In order to tune the hyperparmeters $\lambda_1$ and $\lambda_2$ in Eq.~\eqref{eq:dec1}, we performed grid search of the two hyperparmeters with beam size as 12 on a development dataset. In order to reduce the computational cost, we fixed the hyperparameters for evaluation with all other decoding beams. Since the external LM does not match all the domains in our evaluation set, we only choose a subset from our evaluation set based on the content for the domain adaptation experiments, which has around 2,000 utterances, and are all dictation speech. 

The results are shown in Table~\ref{tab:lm}. When both $\lambda_1$ and $\lambda_2$ are 0, it corresponds to the baseline system without an external LM, and we observe that the WERs vary slightly when using different decoding beams. When $\lambda_1=0$, the method corresponds to the widely used shallow fusion approach~\cite{Hannun2014Deep}, and the results show that this approaches can achieve consistent WER reduction for both NLL and MWER trained models. However, the gain is relatively small, which is only up to 2\% relative. When both $\lambda_1$ and $\lambda_2$ are active, the method corresponds to the default HAT model evaluation with an external LM. The results show that this approach can achieve over 8\% relative WER reduction compared with the baseline system, and over 5\% relative WER reduction compared with the shallow fusion approach. Across all the three decoding conditions, the MWER trained model resulted in lower WER compared with the NLL trained model. For this particular domain, the MWER model may be further improved by using the external LM to generate the N-best list, which will be investigated in our future work.

\begin{table}[t]\centering
\caption{Results from evaluation with the external LM following Eq. \eqref{eq:dec1}. When both $\lambda_1$ and $\lambda_2$ are 0, it corresponds to the baseline results without the external LM. When only $\lambda_1 = 0$, the method corresponds to the shallow fusion approach. }
\label{tab:lm}
\footnotesize
\vskip-2mm
\begin{tabular}{l|c|cccc}
\hline 

\hline
& & \multicolumn{4}{c}{Decoding Beam} \\ 
Model      & LM weights & 4 & 8 & 12 & 16  \\ \hline
 & $(0, 0)$ & 15.9 & 15.5  & 15.5 & 15.6\\
 NLL & (0, $\lambda_2$) & 15.7 & 15.1 & 15.1 & 15.2\\
&  ($\lambda_1$, $\lambda_2$) & 14.7 & 14.3 & 14.3 & 14.3 \\ \hline
 & $(0, 0)$ & 15.6 & 15.4  & 15.3 & 15.4\\
  MWER & (0, $\lambda_2$) & 15.3 & 14.9 & 14.9 & 15.0 \\
&  ($\lambda_1$, $\lambda_2$) & 14.7 & 14.2 & 14.1 & 14.2 \\

\hline

\hline
\end{tabular}
\vskip-4mm
\end{table}

\section{Conclusion}
\label{sec:conc}

In this work, we presented the MWER training of the recently proposed HAT model for E2E speech recognition. From our results, HAT is more sensitive to length normalization during inference compared to the standard RNN-T, and MWER training can narrow down the gap between the reults with or without length normalization. In addition, we also showed that MWER training can improve the robustness of the model with different decoding beam sizes during inference. In terms of LM fusion, our results show that the HAT approach that integrates both the internal and the external LM during scoring works significantly better than the shallow fusion approach when the internal LM component is disabled, while MWER training can further improve the accuracy. 


\bibliographystyle{IEEEbib}
\bibliography{bibtex}

\end{document}